\documentclass[12pt]{article}

\usepackage{scicite}
\usepackage{graphicx}
\usepackage{amsmath}
\usepackage{times}

\newenvironment{sciabstract}{%
\begin{quote} \bf}
{\end{quote}}

\newcounter{lastnote}

\title{Beyond imitation: Zero-shot task transfer on robots by learning
concepts as cognitive programs}

\author
{Miguel L\'azaro-Gredilla$^\ast$, Dianhuan Lin, J. Swaroop Guntupalli, Dileep George$^\ast$\\
\\
\normalsize{Vicarious AI,}\\
\normalsize{California, USA}\\
\\
\normalsize{$^\ast$To whom correspondence should be addressed; E-mail:  \{miguel, dileep\}@vicarious.com.}
}

\date{}

\begin{document}


%\baselineskip24pt


\maketitle


\begin{sciabstract}
Humans can infer concepts from image pairs and apply those in the physical world in a completely different setting, enabling tasks like IKEA assembly from diagrams. If robots could represent and infer high-level concepts, it would significantly improve their ability to understand our intent and to transfer tasks between different environments. To that end, we introduce a computational framework that replicates aspects of human concept learning. Concepts are represented as programs on a novel computer architecture consisting of a visual perception system, working memory, and action controller. The instruction set of this ‘cognitive computer’ has commands for parsing a visual scene, directing gaze and attention, imagining new objects, manipulating the contents of a visual working memory, and controlling arm movement. Inferring a concept corresponds to inducing a program that can transform the input to the output. Some concepts require the use of imagination and recursion. Previously learned concepts simplify the learning of subsequent more elaborate concepts, and create a hierarchy of abstractions. We demonstrate how a robot can use these abstractions to interpret novel concepts presented to it as schematic images, and then apply those concepts in dramatically different situations. By bringing cognitive science ideas on mental imagery, perceptual symbols, embodied cognition, and deictic mechanisms into the realm of machine learning, our work brings us closer to the goal of building robots that have interpretable representations and commonsense.

\end{sciabstract}

% In setting up this template for *Science* papers, we've used both
% the \section* command and the \paragraph* command for topical
% divisions.  Which you use will of course depend on the type of paper
% you're writing.  Review Articles tend to have displayed headings, for
% which \section* is more appropriate; Research Articles, when they have
% formal topical divisions at all, tend to signal them with bold text
% that runs into the paragraph, for which \paragraph* is the right
% choice.  Either way, use the asterisk (*) modifier, as shown, to

\section*{Introduction}

\begin{figure}
    \centering
    \includegraphics[width=5.5in]{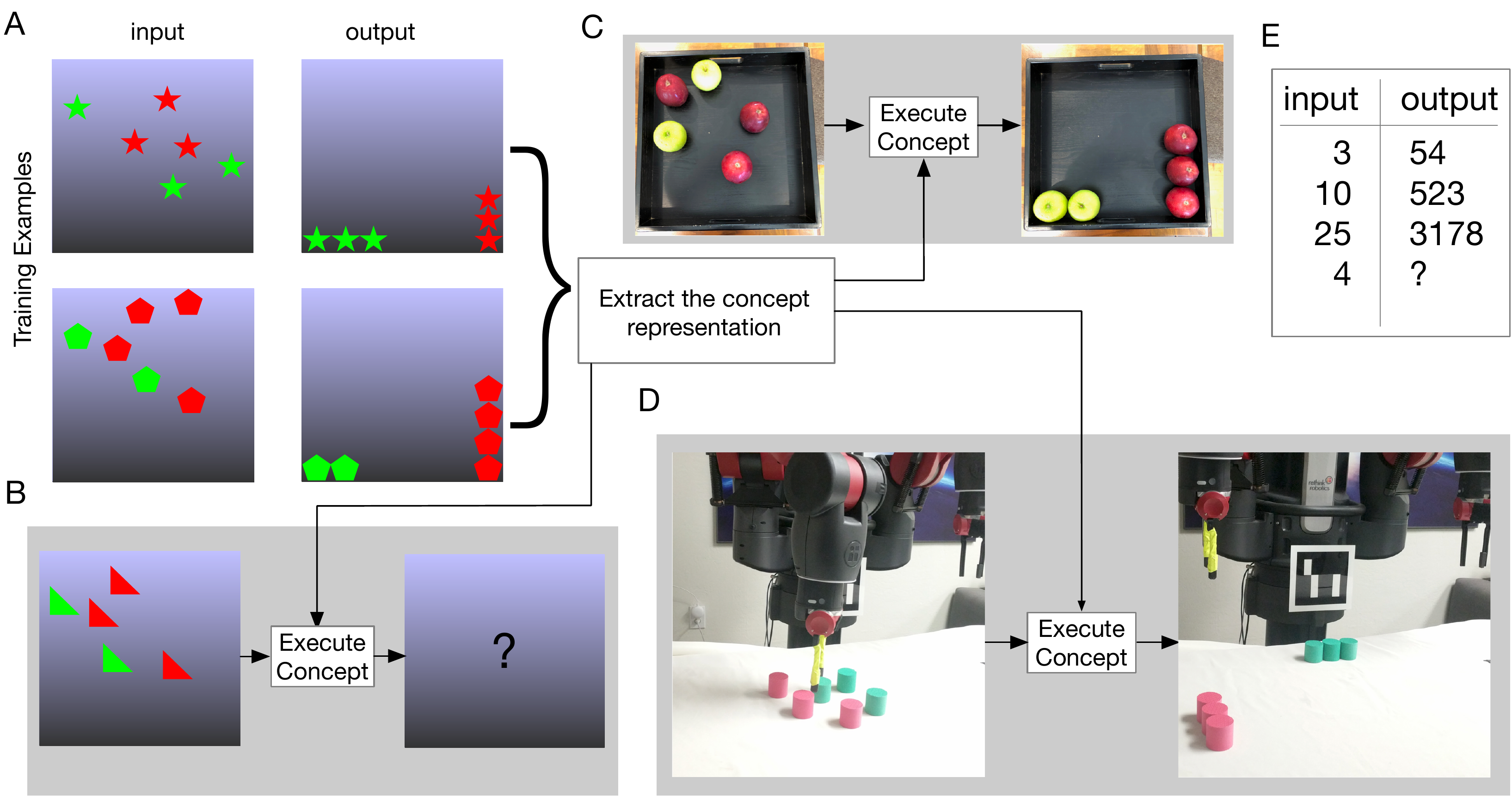}
    \caption{People can easily understand the concept conveyed in pairs of images, a capability that is exploited by LEGO and IKEA assembly diagrams. (A) People interpret the concept conveyed by these images as stacking red objects vertically on the right and green objects horizontally at the bottom. (B) Given a novel image people can predict what the result of executing the concept wold be. (C) Concepts inferred from schematic images as in A can be applied in real-world settings. (D) Enabling robots to understand concepts conveyed in image-pairs will significantly simplify communicating tasks to robots. (E) Not all concepts conveyed as input-output pairs are as readily apparent to humans as the visual and spatial reasoning tasks.}
    \label{fig:1}
\end{figure}

Humans are good at inferring the concepts conveyed in a pair of images and then applying them in a completely different setting, for example the concept of stacking red and green objects in Fig \ref{fig:1}(A), applied to the different settings in Fig \ref{fig:1}(B-D). The human inferred concepts are at a sufficiently high level to be effortlessly applied in situations that look very different, a capacity so natural that it is utilized by IKEA and LEGO to make language-independent assembly instructions. In contrast, currently robots are programmed by tediously specifying  the desired object locations and poses, or by imitation learning where the robot mimics the actions from a demonstration \cite{Akgun2012,DuanOne-shotLearning, HuangNeuralDemonstration, FinnOne-shotMeta-learning}. By relying on brittle stimulus-response mapping from image frames to actions, the imitation-learning policies often do not generalize to variations in the environment which might include changes in size, shape and appearance of objects, their relative positions, background clutter, and lighting conditions \cite{TungRewardDemonstrations}.

If, like people, a robot could extract the conceptual representation from pairs of images given as training examples (Fig \ref{fig:1} A), and then apply the concept in dramatically different situations and embodiments, it would greatly increase their flexibility in adapting to new situations and to unstructured environments. A shared conceptual structure with humans would also simplify communicating tasks to a robot at a high-level, and help to improve their conceptual repertoire with further interactions. 

A concept is a re-description of everyday experience into a higher level of abstraction \cite{Barsalou1999, Mandler1992}. One way to characterize the pictures in Fig \ref{fig:1} is a pixel-by-pixel description of the changes from the input image to the output, a description that will not generalize to new situations. 
Concepts enable a higher level of description that generalizes to new situations, and ground \cite{Cangelosi2002SymbolHypothesis,PhenomenaTheProblem} verbal expressions like ``stack green objects on the right'' with real-world referents. In contrast to the visuo-spatial concepts like the one in Fig \ref{fig:1}(A) that are easy and immediate even for children \cite{Amalric2017ThePreschoolers}, numerical concepts like the one shown in \ref{fig:1}(E) are neither easy nor immediate for people. The concepts that are easy and immediate, and form the basis of commonsense in humans, are a very small subset of all  potential concepts.

In this paper we hypothesize that human concepts are programs, termed cognitive programs \cite{Tsotsos2014},  implemented on a computer with certain constraints and architectural biases (eg. a {\em 'Human Turing Machine'} \cite{Zylberberg2011}) that are fundamentally different compared to the prevalent von Neumann architecture. Under this hypothesis, the inductive biases encoded in the architecture and instruction set of this computer explain why visuo-spatial concepts like those in Fig \ref{fig:1}(A), are easy and intuitive for humans whereas the numeric concept shown in Fig \ref{fig:1}(E) is harder and non-intuitive. 
In this view, concepts are sequencing of elemental operations \cite{Roelfsema2005} on a cognitive computer according to a  probabilistic language of thought \cite{Yildirim2015LearningApproach, Piantadosi2016}, and their generalization to new situations arise out of skillful deployment of visual attention, imagination, and actions.

Our current work builds upon and is related to several key ideas from both cognitive and systems neuroscience, such as visual routines \cite{Ullman1996} forming the basis for high-level thinking \cite{Roelfsema2005}, perceptual symbol systems \cite{Barsalou1999}, image schemas \cite{JohnsonInferringReasoning, Mandler1992, MandlerOnSchemas}, deictic mechanisms \cite{ballard1991animate}, mental imagery and the role of V1 as blackboard for mental manipulation \cite{Roelfsema2016}, and brings them into the foray of machine learning. Following the ideas of perceptual symbol systems \cite{Barsalou1999} and image schemas \cite{LakoffWhereBeing}, we treat concepts as simulations in a sensori-motor system with imageable spatial information forming its fundamental building block \cite{MandlerOnSchemas}. The main components of our architecture include a vision hierarchy \cite{George2017}, a dynamics model for interactions between objects \cite{kansky2017schema}, an attention controller, an imagination blackboard, a limb controller, a set of primitives, and program induction. 
%Our proposed architecture is fundamentally different from the prevalent von Neumann architecture or any other Turing machine. 
We evaluate our architecture on its ability to represent and infer visual and spatial concepts that cognitive scientists consider to be the fundamental building blocks \cite{MandlerOnSchemas}. By building a working computational model and by evaluating it on real-world robotics applications, we bring several of these ideas which exist purely as descriptive theories into a concrete framework useful for hypothesis testing and applications. 

Given input-output examples indicating a concept, we induce programs in this architecture, execute those programs in new settings, and on robots to perform desired tasks (Fig \ref{fig:1}B-D). In contrast to imitation learning where a robot mimics a demonstration in the same setting, we show that the cognitive programs induced on our proposed architecture generalize well to dramatically new settings without explicit demonstrations. Cognitive programs exhibit schematic-to-real transfer similar to the capability of humans to understand concepts from schematic drawings to apply them in situations that look very different \cite{lake2016building}.

\section*{Cognitive programs on a visual cognitive computer}

The Visual Cognitive Computer (VCC) is a mechanistic integration of visual perception, a dynamics model, actions, attention control, working memories and deictic mechanisms into a computer architecture.
The design process for VCC involved starting with architectural sketches provided in previous works \cite{Tsotsos2014,Zylberberg2011} based on cognitive- \cite{Ullman1996,Barsalou1999} and neuro- \cite{Roelfsema2005,Roelfsema2016} science considerations.
These architectural sketches provided functional requirements, rough block diagrams, and descriptive theories as a starting point, but no computational models.
This initial architecture is then refined and extended by co-designing it with an instruction set from the viewpoint of succinctness and completeness in program induction \cite{GulwaniInductiveWorld, OverlanLearningThought}. The design elements are finalized when an appropriate instruction set can be executed on the machine with adequate support for program learning.

\begin{figure}
    \centering
    \includegraphics[width=6.2in]{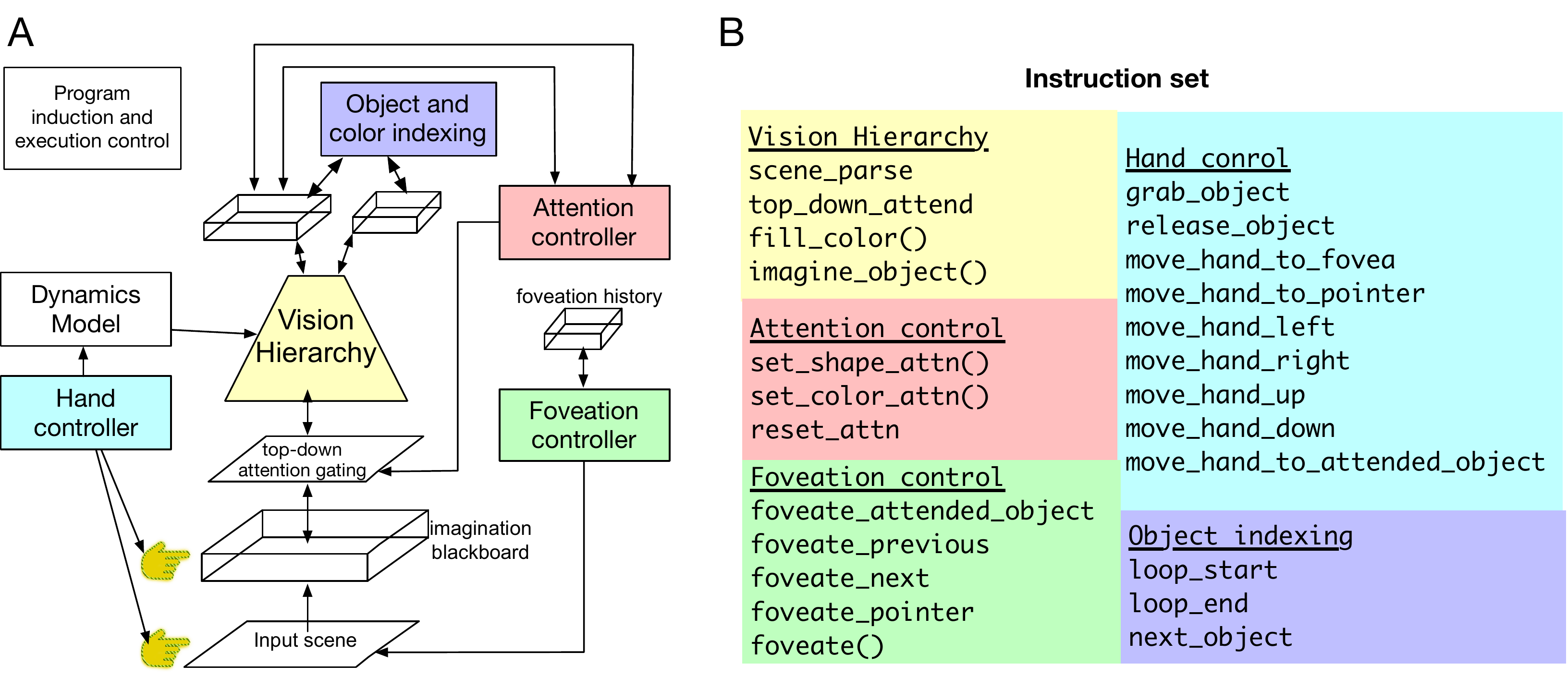}
    \caption{Architecture (A) and instruction set (B) of the visual cognitive computer. Vision hierarchy parses an input scene into obejcts, and can attend to objects and imagine them. Hand controller has commands for moving the hand to different locations in the scene, and fovea controller for positioning the fovea. Object indexing commands iterate through the objects currently attended to. Attention controller can set the current attention based on object shape or color. }
    \label{fig:2}
\end{figure}

Figure \ref{fig:2}(A) shows the architecture of VCC which includes the embodiment of the agent. The agent has an eye whose fovea (the center of the visual field) can be positioned within the input scene, and a hand that can be used to move objects either in the real world or in imagination. The hand controller positions the hand in the visual scene and grabs and releases objects, and foveation controller changes the location of the fovea in the scene. The vision hierarchy (VH) can parse input scenes containing multiple objects and imagine objects, similar to the generative model we developed in \cite{George2017}. Parsed objects of a scene are stored in the object-indexing memory, and they are ordered according to their distance from the current location of the fovea. An important component of the architecture is the imagination blackboard to which objects can be written and manipulated. The dynamics model combined with the vision hierarchy lets the system predict the effect of imagined movements, to write those results into the imagination blackboard.  The attention controller is used to selectively and in a top-down manner attend to objects based on their category or color. Top-down attention also acts as an internal pointing mechanism \cite{Ballard1997} to reference objects in the imagination blackboard. An external agent, a teacher for instance, can interact with the VCC agent by showing it images and by directing its attention using a pointer \cite{Tomasello1999}.

In addition to the imagination blackboard, VCC has other structured local working memories for object indexing, color indexing, and foveation history. The working memories are structured in how they represent their content, and the locality of the working memory to their controllers enforces structured access; the instructions that can read from and write to specific memories are pre-specified. Fig \ref{fig:2}(B) lists the instruction set of VCC, and their mapping to the different controllers.
The instruction set of the VCC was heavily influenced by the primacy of objects and spatial primitives in human cognition \cite{MacbethImageComparison}, and by the elemental operations that have been previously proposed \cite{Tsotsos2014,Roelfsema2005}. Looping constructs {\tt loop\_start, loop\_end} are constrained to loop over the currently attended objects in a scene. See the supplementary material for implementation details of the instruction set.

One critical design consideration for VCC is the ease of program induction. As an effect of having working memories that are local and specific to instructions rather than as generic registers in a von Neumann architecture, the program induction search is vastly simplified due to the fewer unbound variables in a candidate program (see supplementary Fig S1 for input and output working memory mappings of some of the instructions). The instructions {\tt set\_color\_attn, set\_shape\_attn, foveate,} and
{\tt imagine\_object} have arguments that determine their output.
During program induction, the arguments to be used for a particular input-output pair can be
explicitly searched, or predicted from the inputs using neural networks that are trained for that purpose. In addition, the arguments to the {\tt foveate} command can be set externally by a teacher by the use of a pointer. A learner that takes advantage of `foveation guidance' from a teacher, or accurate predictions of instruction-arguments from a neural network can speed up induction by reducing the search space.

\begin{figure}
    \centering
    \includegraphics[width=6in]{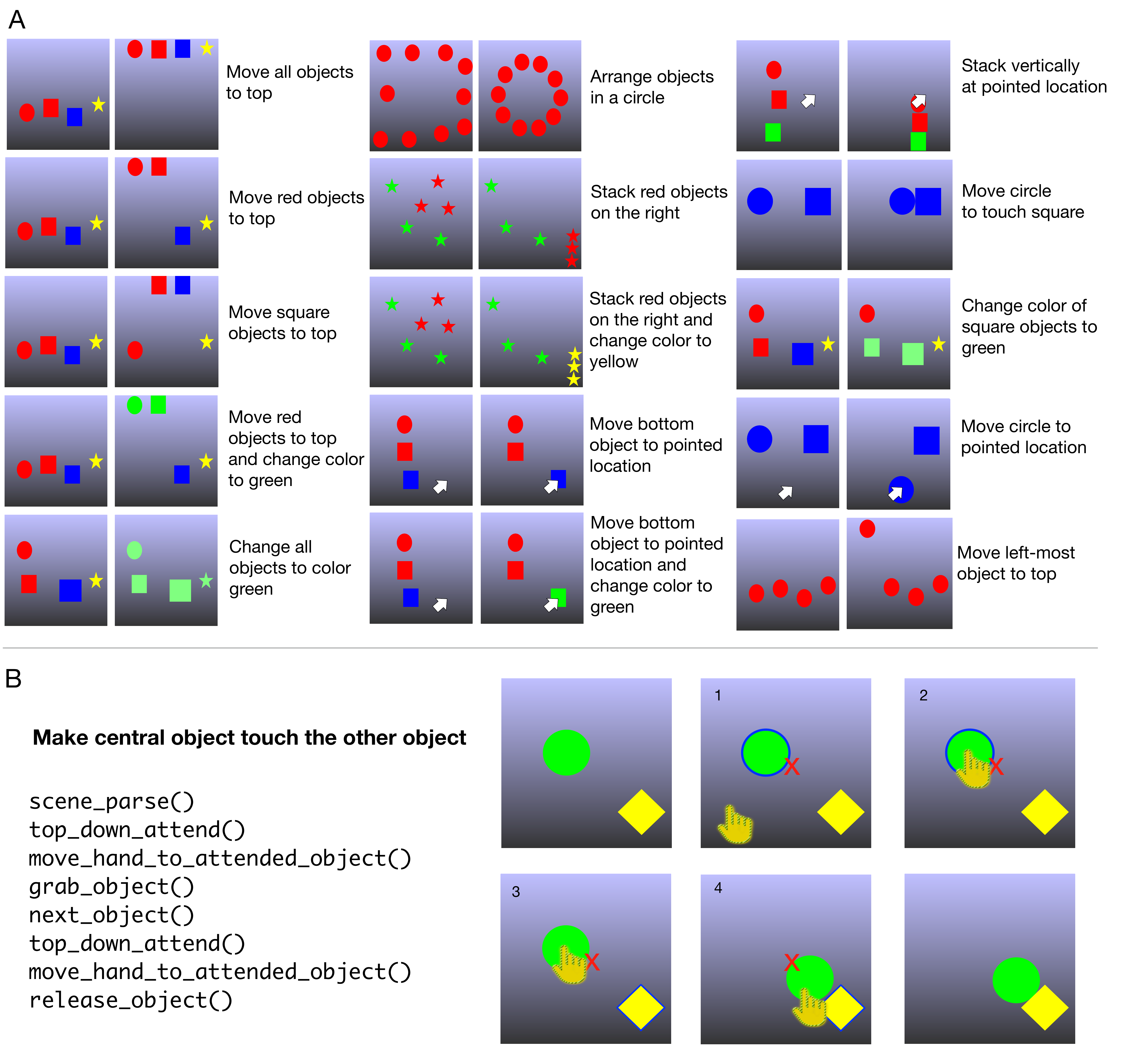}
    \caption{Concepts and their representation as cognitive programs. (A) Input-output examples for 15 different tabletop concepts. In our work we test on 546 different concepts (see supplementary materials for the full list). (B) Manually written program for a concept that requires moving the central object to touch the other object. The panels on the right show different stages during the execution of the program. The attended object is indicated by a blue outline.}
    \label{fig:3}
\end{figure}

To guide the design and evaluation of VCC, we use visual concepts in `tabletop world' (TW) corresponding to the spatial arrangement of objects on a tabletop Fig \ref{fig:3}(A). TW allows us to focus our investigation on imageable objects and spatial primitives that are considered to be the first conceptual building blocks \cite{MandlerOnSchemas}.
Tabletop world has the advantage of having a rich set of concepts to rigorously test concept representation and generalization, while having simplified dynamics and perceptual requirements compared to the full complexity of the real world. Object motions are limited to sliding along the surface of the table and they stop moving when they collide with other objects or with the boundary of the workspace. How an agent generalizes its acquired concepts depend on the regularities of the world it is exposed to, and on the regularities in its interaction with the world. By being a proper subset of the real world, TW 
enables representation and discovery of concepts from schematic images while still having real-world counterparts. The infinite number of physical realizations of each tabletop world concept also enables testing for strong generalization on a physical robot and for transfer from schematic diagrams to real-world.

Figure \ref{fig:3}(B) shows a simple, manually written cognitive program that represents a concept, and serves to illustrate the representational properties of the VCC. The concept is about making the object close to the center touch the other object in the scene. Since the fovea is centered by default, the centered object is highlighted  on the first call to {\tt top\_down\_attend}. After moving the hand towards that object and grabbing it, {\tt next\_object} command is used to switch the attention to the other object. Moving the grabbed object towards the other object until collision achieves the desired goal of making the objects touch. Stopping the object when it comes into contact with another object requires access to the details of the outer contours of the object, a detail that is available only at the bottom of the visual hierarchy. In contrast to architectures like auto-encoders that map a visual input to a fixed encoding in a feed-forward manner \cite{kingma2013auto}, the imagination buffer of the VCC allows for the details of the object to be represented and accessed from the real world as required. As anticipated in \cite{Barsalou1999} regarding the requirements of a perceptual symbol system, this allows the vision hierarchy to be part of an interactive querying of the world as part of a program rather than trying to solve the problem of representing a scene in one feed-forward pass for downstream tasks. In our current work, we assume that the vision hierarchy (VH) and dynamics are already learned as described in our earlier works \cite{George2017,kansky2017schema}.

The problem of learning to represent concepts is this: given a set of input-output image pairs representing a concept, induce a cognitive program that will produce the correct output image when executed on the VCC with the corresponding input image. To solve this problem, we combine recent advances in program induction \cite{Gulwani2010,BalogDeepcoder:Programs, ChenTOWARDSEXAMPLES,Dechter2013,GravesNTM,Lin2014, Ellis2018DREAMCODER}
 along with the architectural advantages of VCC.
 Our overall approach can be understood in the explore-compress (E-C) framework \cite{Dechter2013} for program induction where induction proceeds by alternating between an exploration phase and a compression phase. During the exploration phase an existing generative model for the space of programs is used to guide the exploration of new programs, and during the compression phase all the programs found so far are used to update the generative model for the space of programs. The generative model for programs works as an input-agnostic prior for the space of future programs given the concepts learned earlier. We use Markov models of instruction-to-instruction transitions and sub-routines that replace a sequence of instructions with a new atomic instruction as the input-agnostic generative models. In the exploration phase, this prior is combined with `argument predictions', predictions from neural networks \cite{BalogDeepcoder:Programs} about the value that the argument of each instruction will take, given that the instruction is part of the program. This prediction is conditional on the specific input-output pairs that the program is being induced for. In addition, the run-time exceptions generated from the VCC are used to prune the search space. In this regard the VCC can be thought of as an `embodiment' that disallows certain instruction-argument combinations during execution, just like the physical attributes of a body restricts some kinds of movements. Although the VCC instructions are named to be meaningful to humans to help with interpretation, the agent is unaware of the meanings of these actions or how the different working memories and indexing mechanisms can be leveraged to represent concepts. By learning programs the agent has to discover how to exploit the actions and properties of the VCC to represent various concepts.

\section*{Results}

\begin{figure}
    \centering
    \includegraphics[width=6in]{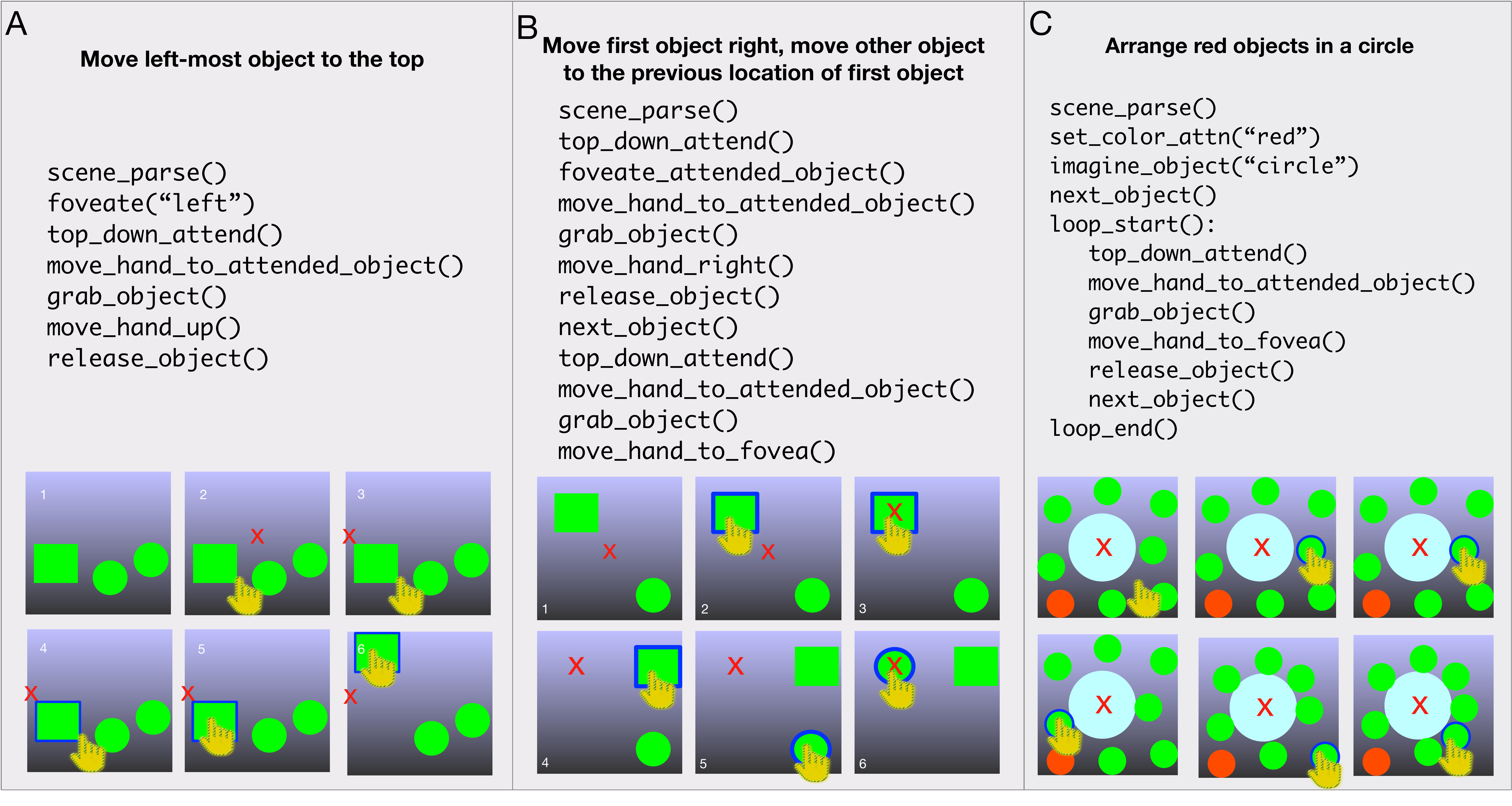}
    \caption{Three example programs from the list of 535 induced programs. These programs demonstrate how the induction method learns to utilize the properties of the VCC to represent different concepts. (A) Moving the left-most object utilizes the ordering of objects relative to the fixation point. (B) Program learning discovered that the position of the fovea can be used as a deictic mechanism and as a working memory. After the object is attended to and grabbed, fovea is moved to the location of that object. When the object is moved away, the location of the fovea maintains the memory of its initial position. (C) Arranging small circles into a large circle is achieved by first imagining the large circle, which doesn't exist in the input image or in the output image, and then pushing the other objects toward it.}
    \label{fig:three_concepts}
\end{figure}

\begin{figure}
    \centering
    \includegraphics[width=5.5in]{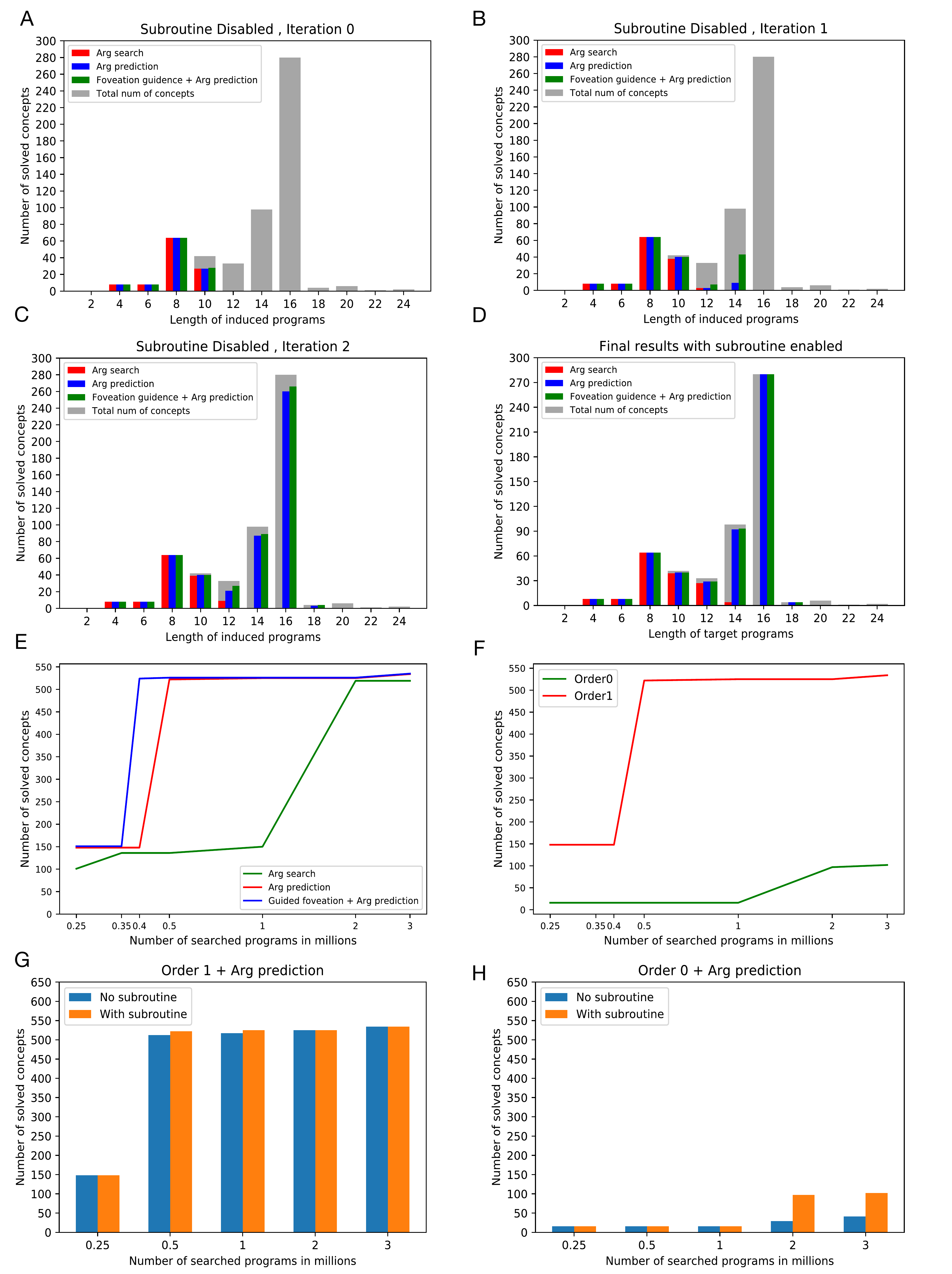}
    \caption{Program induction details. (A-C) Length distribution of induced programs for the first three E-C iterations. X-axis bins correspond to program lengths (number of atomic instructions). The gray bars represent the total number of programs of that length according to a set of manually written programs comprising all concepts. (D) Distribution at the end of all iterations. (E,F) Number of induced programs for different search budgets, for different model options. (G,H) Effect of subroutines.}
    \label{fig:iterations}
\end{figure}

\begin{figure}
    \centering
    \includegraphics[width=5.5in]{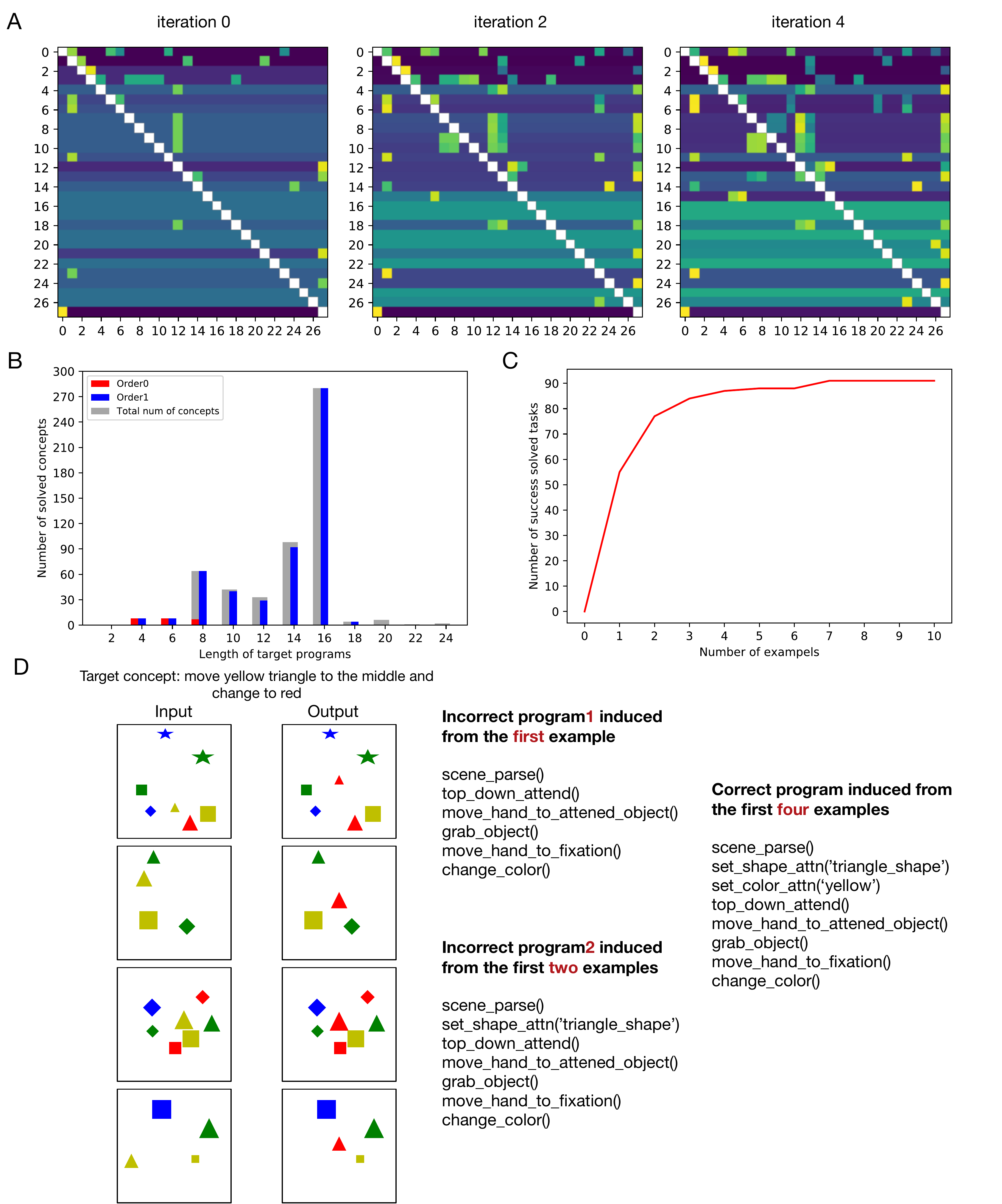}
    \caption{ Program induction and generalization. (A) The instruction-to-instruction transition matrix after different E-C iterations. (B) Length distribution of programs induced using order-0 vs order-1 model. (C) Training curve. Most concepts are solved with just a few examples. (D) An example showing wrongly induced programs when only three training examples from a concept are presented, where accidental patterns in the data can explain the examples. In this case the correct concept was induced with 4 examples.}
    \label{fig:5}
\end{figure}

Our experiments show that concepts can be induced as cognitive programs and that this enables transfer from diagrammatic representations to execution on a robot without any explicit demonstration. We evaluated the performance of VCC on 546 different concepts, for which manually written programs vary in lengths from 4 to 23  (Fig \ref{fig:three_concepts}). We investigated how model-based search, input-conditional argument prediction, and foveation guidance by a teacher can affect how quickly the programs can be learned. Using a combination of the best models, argument prediction, and foveation guidance, 535 of the 546 concepts could be learned with a search budget of 3 million programs, and 526 of them could be learned with a search budget of 0.5 million programs.
The induced concepts readily generalize to settings where we vary the number and size of objects, their appearance, and the color and texture of the background. They can also be readily transferred to novel situations on multiple robots to execute tasks in the real world.

\subsection*{Induced programs}
Figure \ref{fig:three_concepts} shows three examples from the 535 induced programs and these demonstrate how the properties of the VCC were utilized by program induction to represent concepts. The first example (Fig \ref{fig:three_concepts} A) is the concept requiring to move the left-most object to the top. The learned program utilizes the fact that objects in the object-indexing memory of VCC are ordered by their distance from the center of fixation. Although left-most is not an attribute that is directly encoded in VCC, that concept is expressed by combining other primitive actions in the VCC.

Figure \ref{fig:three_concepts}(B) shows an example where induction discovered the use of a deictic mechanism. The concept required moving the central object to the right edge and moving the other object to the position previously occupied by first object. Achieving this requires holding the position of the previous object in the working memory. The induction method discovered that the fovea can be used as a pointer to index and hold object locations in memory. The discovered program moves the fovea to the first object after attending to it and grabbing it. Since the fovea remains at the initial position of the first object when it is moved away, this allows for the second object to be moved to the correct position by simply moving it to the fovea. The concept of arranging small circles into a big circle (Fig \ref{fig:three_concepts} C) requires the agent to imagine an object that doesn't exist (the big circle), and then push other objects toward this imagined object while it is maintained in the working memory. This is also an example where teaching by demonstration is likely to face challenges because the intermediate stages of a demonstration would provide only scant cues about the final goal of arranging objects in a circle.

\subsection*{Induction dynamics}

We tested the efficacy of program induction under a wide variety of settings. For model based search, we compared using an order-0 model that only considers the relative frequencies of instructions to an order-1 model that learns the transition probabilities of instructions in programs. In each case, we tested the effect of learning subroutines.  We tested the effect of combining the order-1 model with neural network-based predictions for the arguments of each instruction. We also tested the effect of explicit foveation guidance where the pointing action by a teacher can be used to set the foveation location in the learning agent.

Iteratively discovering the programs by modifying the model that is driving the search process, and augmenting that model with input-conditioned argument predictions are both important for discovering the complex concepts with longer programs. Figure \ref{fig:iterations}(A-D) shows the length distributions of the programs induced after each explore-compress iteration for different settings, for a search budget of one million programs. The histogram of the number of programs of different lengths are plotted along with a ground-truth  distribution. (The ground-truth is derived based on manually writing programs for each of the concepts. The discovered program for any particular concept need not match the ground truth program in implementation or length). When argument prediction is not available, program induction relies on searching exhaustively in the argument space. Without argument prediction, no programs of length more than 12 are induced, whereas with input-conditioned prediction of arguments, programs up to length 18 are induced. In all, without argument prediction only 150 of the 546 concepts are induced, whereas argument prediction and subroutines enable the discovery of 526 out of 546 concepts. Iterative refinement of the model using the discovered programs at each stage results in more programs being discovered in the next iteration of search. Figure \ref{fig:iterations}(A-C) show the length distributions of programs discovered after each explore-compress iteration. The distributions of the discovered programs shift right with each iteration, successively discovering longer programs that were missed in the previous iteration. The progression of concepts learned during multiple E-C iterations demonstrate the advantage of learning to learn \cite{Lake2015} in program induction.

A teacher can help the agent learn faster by guiding the agent's attention using a pointer. We tested the effect of having a teacher guide the foveation of the agent for concepts that require using the {\tt foveate()} function similar to the joint attention mechanism hypothesized in cognitive science \cite{Tomasello1999}. This was achieved by setting the argument of the {\tt foveate()} function to the location pointed by the teacher. In contrast with other instructions, the arguments for the {\tt foveate()} function are not predictable from the  input-output image-pairs, and program induction relies on searching the argument space in the absence of any other guidance. The effect of offering foveation is seen in Fig \ref{fig:iterations}(E) as the search budget is reduced from 1 million programs to 0.4 million. In this setting using foveation guidance increases the number of discovered programs from 148 to 524. Although foveation guidance from a teacher was applied here in a very limited setting, this shows that pre-specified joint attention mechanisms with a teacher hold promise in being able to significantly speed up the learning.

Overall, using model-based search using an order-1 model and argument prediction were the most important factors that determined how quickly the programs could be learned. Figure \ref{fig:iterations}(E-H) show the effect of various factors as the search budget is varied from 0.25 million programs to 3 million programs. When an order-1 model was used, induction of subroutines played a smaller role in the induction of new programs: while subroutines that reduced the description length were obtained after each compression iteration, they provided only a modest help in the future discovery of programs. To give a sense of the required computational effort, searching 1 million programs took around 10 minutes when using a single core.

Modeling the sequential relationship between the instructions of learned concepts significantly helps with program induction, compared to modeling just the instruction frequencies. We tested an order-0 model that ignores the sequential dependencies, and an order-1 Markov model between the instructions as the models used in the E-C iterations.
Out of 546 concepts, and with a search budget of 1 million programs, the order-0 model was not able to induce any new concept when subroutines were disabled, and only 7 new concepts when they were enabled\footnote{We also tested the effect of increasing the search budget. When increasing it to 3 million programs, 41 concepts are discovered without subroutines and an additional 61 when using subroutines. The effect of subroutines in concept discovery is much more marked in the case of the order-0 model, because subroutines are the only mechanism that makes memory available to the model.}. In contrast, an order-1 model was able to learn 525 of the 546 concepts for the same search budget (Fig \ref{fig:5} B). Figure \ref{fig:5}(A) shows the transition matrix of the order-1 Markov model after iterations 0, 2, and 4. The transition matrix learns additional relevant modifications with each iteration that enables more concepts to be learned in future iterations.

\begin{figure}
    \centering
    \includegraphics[width=4.5in]{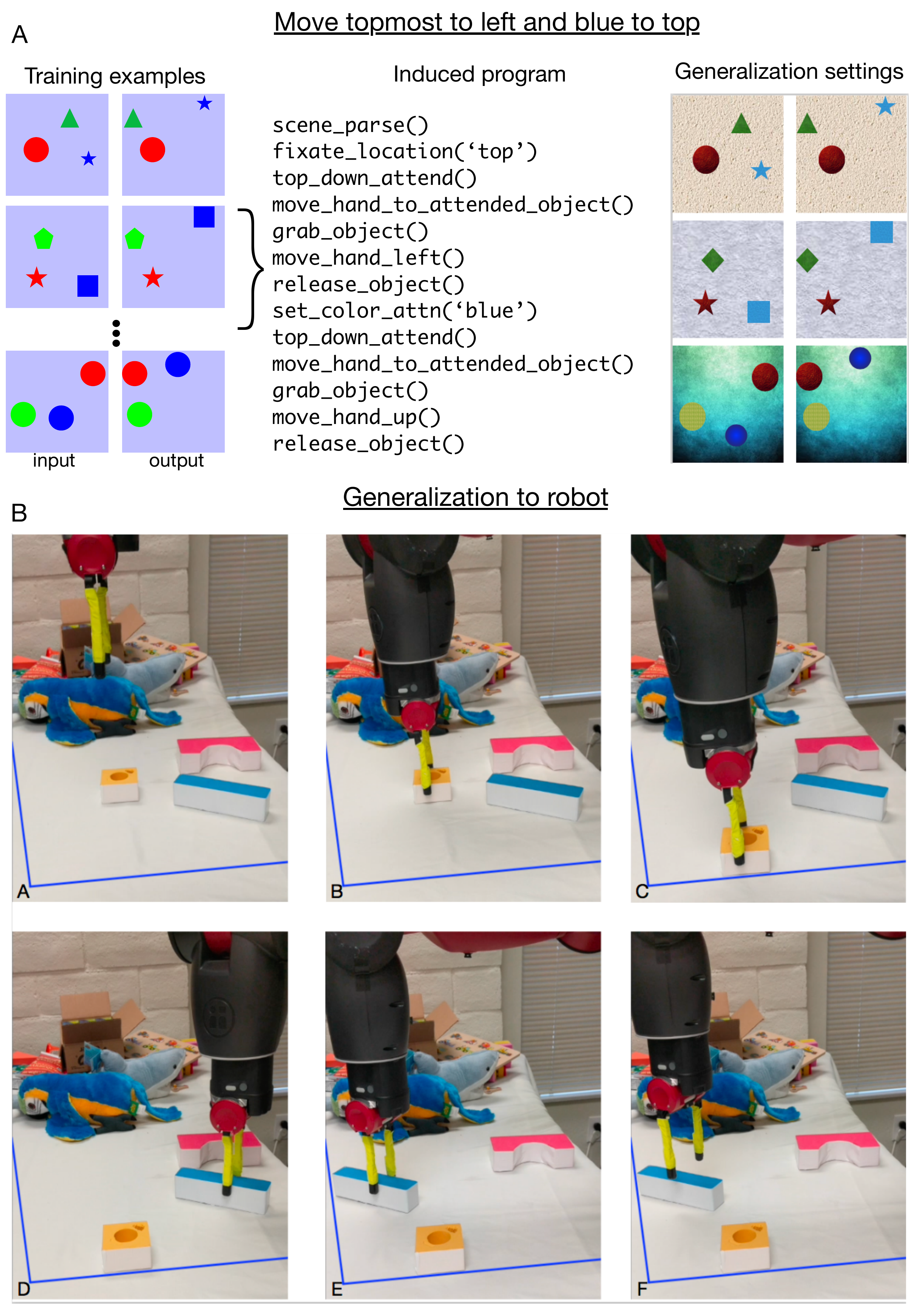}
    \caption{Generalizing to new settings and to the real world. (A) Training examples and induced program corresponding to the concept 'Move topmost to left and blue to top'. The right column shows different test settings to which the program generalizes. The test settings are all very different from the training setting except for their conceptual content. (B) The concept in A executed on Baxter robot with very different objects compared to the training setting. Different stages in the execution are shown.}
    \label{fig:6}
\end{figure}

Most concepts were learned correctly with just a few input-output examples, and generalized well to dramatically different images depicting the same concepts. Figure \ref{fig:5}(C) shows the learning curve for concepts discovered in the first E-C iteration (107 concepts, including 16 bootstrapping ones). The learning curve flattens out after 5 examples. Figure \ref{fig:5}(D) shows an example of the confusion caused when the number of examples are fewer. While the concept required moving the black triangle to the middle, in the first two examples this could be achieved just by moving to the middle whichever triangle was closest to the center.
The generalization of the induced concepts to new situations was tested using concept images generated with different numbers of objects and visual transformations, examples of which are shown in Fig \ref{fig:6}(A). The test images were generated by varying the backgrounds, object-foregrounds, number and kinds of objects and their placements within the image.  For each concept, we tested on 50 different test images generated by sampling from these parameters. More examples of test images are shown in supplementary Figure S4. All the induced concepts generalized to a 100\% of these new settings.

\begin{figure}
    \centering
    \includegraphics[width=5.5in]{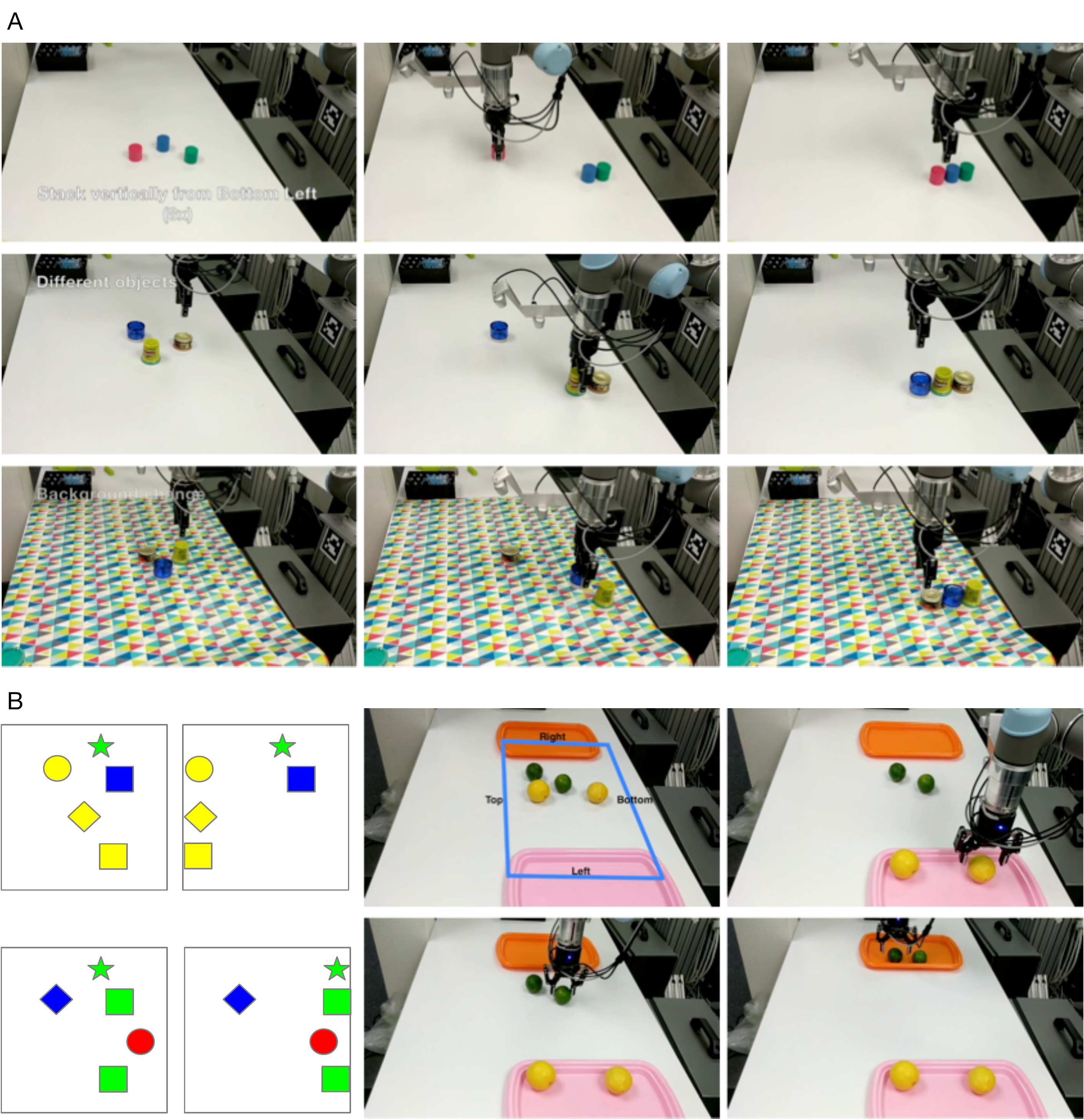}
    \caption{Learned concepts transferred to different real-world settings. (A) Each row shows the starting state, an intermediate state, and the ending state for three different execution scenarios for a concept that requires stacking objects on the bottom left. Middle row shows execution on different objects and the bottom row shows execution on a different background. (B) Execution frames from an application that separates limes from lemons. This task is achieved by the sequential composition of two concepts. Left panel shows the two concepts used (top and bottom) and right panel shows execution of these concepts in sequence to achieve the task.}
    \label{fig:robot-exec}
\end{figure}

\subsection*{Transfer to robots}

Induced concepts transferred readily from the schematic inputs to the real world execution on two different robots: a Baxter from Rethink Robotics and an UR5 from Universal Robots. The primitive instructions for perception and action, such as capturing the input scene and controlling hand, were implemented using the respective robots' interface while the rest of the VCC instructions remained the same. We tested the transfer of 5 different induced concepts on the two robots, and each concept was tested in several different scenarios by changing the number and types of objects, their positions, the number of distractor objects, and the appearance of the backgrounds. In each case, the initial state of the test scenarios were constructed such that running the induced program on the input image would produce the right output in VCC's imagination.

Figures \ref{fig:6} and \ref{fig:robot-exec} show the robots at different stages of the execution of the concepts for a sampling of the settings we tested on. The concepts are executed correctly despite the variations as long as the information relevant for the concept is present in the input, invariant to the number of objects, their sizes, positions and variations in the background. In Fig \ref{fig:robot-exec}(A), we show the stacking concept executed in three different settings. Notably, the different stacking concepts are  executed correctly with the objects properly in contact in their final states, even when the sizes and types of objects are significantly different. Out of the total  test scenarios that were tried, the robot UR5 succeeded in executing the concepts in more than 90\% of the trials. The failures were caused by the objects slipping out of the gripper during a grasp, or during placement. On Baxter the success rate was lower (at about 70\%) due to blurry camera, less effective gripper and imprecise motion in our aging robot.

We also show how a complex concept that couldn't be induced directly could be broken into two different concepts and executed consecutively on the robot to achieve the more complex task. The final task requires moving the yellow objects on the table to the left and the green object on the table to the right, which can be achieved as a combination of the first set of movements with the second. We show an application where the concept can be used to separate lemons from limes (Fig \ref{fig:robot-exec}B). The reader is encouraged to view the robot demonstration videos available as supplementary material.

\section*{Discussion}

Getting robots to perform tasks without explicit programming is one of the goals in artificial intelligence and robotics. Imitation learning, a predominant approach to this end, seeks to teach the robots via demonstration. To demonstrate the actions required for a task, a robot could be physically guided or operated remotely \cite{Akgun2012, Whitney2017}. Meta-learning approaches try to learn executable policies from a single demonstration \cite{DuanOne-shotLearning, FinnOne-shotMeta-learning, WuTowardsRobots}. The main drawbacks of imitation learning approaches are their focus on rote mimicry without understanding the intent of a demonstration, and their tendency to latch on to surface statistics and idiosyncrasies of the demonstration \cite{JohnsonInferringReasoning} rather than its conceptual content, resulting in limited generalization to new settings. Recent work \cite{HuangNeuralDemonstration} has sought to build an intermediate representation from a demonstration that would allow generalization to more settings, but still relies on having a demonstration available for each variation of a task, in addition to having a large annotated background set for the family of variations. 

By focusing on the discovery of conceptual understanding from pairs of images, our work is very different from the traditional setting of imitation learning. No demonstrations are available, and the agent has to discover the conceptual content and transformation that is represented by the image pairs. Moreover, the discovered representations need to transfer to very different settings, including dramatically different visual appearances and different robot embodiments, from what was shown in the schematic images during training. Such transfer requires visual analogy making \cite{Hofstadter1994TheAnalogy-making., FrenchTabletop:Analogy-Making} as opposed to rote imitation. While we take inspiration from earlier works on learning programs to describe the contents of images and videos \cite{Ganin2018SynthesizingLearning, Tremblay2018SyntheticallyDemonstrations, Lake2015}, those works focused on settings where the content to be described was fully observable in the images or video demonstration. Our setting is very different and significantly more challenging because of the need to discover hidden variables underlying the concepts, the number of steps involved in going from an input image to the output, and because of the need for strong schematic-to-real generalization in settings significantly different from the training distribution.

The sequential and programmatic nature of conceptual representations has been well recognized in cognitive science and neuroscience, with Ullman's seminal paper on visual routines \cite{Ullman1996}, and prevalent cognitive architectures \cite{ANDERSEN2003,Newell1992UnifiedSoar} that use sequential representations. While sharing the motivations of sequential representations and chunking, our work is a significant departure from the traditional cognitive architectures of ACT-R and SOAR in emphasizing the perceptual, motor, and interactive nature of concept representations. In having a vision hierarchy that is not a passive feed-forward transducer, but an active system that can be manipulated and reconfigured \cite{Tsotsos2014}, our architecture follows the guidelines of perceptual symbol systems \cite{Barsalou1999}, rather than pure symbol manipulation. VCC and cognitive programs can also be thought of as a concrete computational instantiation of image schemas \cite{Lawler1983}, an influential idea in cognitive science that emphasizes the grounded and embodied nature of concepts. The programs that we learn on VCC can be considered as `patterns of sensorimotor experience' that constitute a concept, forming the basis for communication and grounded language \cite{Kolodny2018}. Our work is also consistent with the idea of a probabilistic language of thought \cite{Goodman2014ConceptsThought} where complex representations are formed from small repertoire of primitives \cite{OverlanLearningThought,Yildirim2015LearningApproach}. Representing concepts as programs makes them naturally compositional as required for a language of thought \cite{Goodman2014ConceptsThought}, and programs provide explainability that has been lacking in current black box policy learning methods.

The tremendous success of deep learning along with discoveries of its limitations has rekindled the age old debate about the role of innate biases vs tabula rasa learning \cite{arXiv:1801.00631DeepAppraisal}. Bucking the recent trend of tabula rasa learning with large training sets, our work here focused on an learning programs on a biased computer architecture whose biases were inspired by cognitive and neuro sciences and by computational considerations. We carefully considered the question of what is a good architectural starting point \cite{lake2016building} for the learning of concepts. This raises many questions: What should be the design of the processor \cite{Zylberberg2011}? What should be the elemental operations \cite{Roelfsema2016}? We believe that the seeming arbitrariness and ambiguity of the starting point is part of the challenge in bringing these ideas into a computational model, and need to be confronted directly \cite{Marcus2014}. We tackled this challenge by treating this as analogous to the design of a microprocessor
where significant design choices need to be made in the number of registers, memory, and instruction set during the design process. As described earlier, cognitive and neuro sciences provided significant guidance, which were then refined from the view point of program induction. While it is unlikely that all the details of the choices we made in this first iteration are the ones that will give rise to human-like concept representation and acquisition, we believe that many aspects of VCC regarding the interaction of visual hierarchy, imagery, working memory, attention, action, and program induction are likely to remain in future iterations. 
Through further theorizing and experiments, the design of the VCC will be expanded and refined. We believe this iterative refinement of architectural priors, inductive biases and learning algorithms to be an essential part of building systems that work like the human mind \cite{lake2016building}.

A particularly important design choice is that of visual hierarchy and top-down attention. Rather than follow the prevailing machine learning view of treating the visual hierarchy as re-encoding the input for 'down-stream tasks' \cite{Yamins}, we treat the visual hierarchy as a mechanism for structured interaction with the world. In this sense, the encoding of the world is not just at the top of the hierarchy, and all details need not be represented at the top of the visual hierarchy in a loss-less manner as in deep neural network generative models. The VH in our model is intentionally lossy because the details can be accessed at the input on demand, consistent with neuroscience ideas of using the primary visual cortex as a high-resolution buffer for mental imagery \cite{Roelfsema2016}. In this design, the pattern of accessing the detail at the bottom, or the more abstract representation at the top becomes part of the representation of the concept.

Many of the existing datasets that measure conceptual and abstract reasoning have drawbacks that prevent them from being used in a study for acquiring and representing concepts as sensorimotor programs. Raven's progressive matrices (RPM) \cite{raven1938raven} are often used as a test for conceptual representation. Instead of using RPM, we chose to use tabletop world (TW) because the properties of the world that gives rise to generalization are systematic and well understood. In contrast, in RPMs, the source of generalization can encompass the full experience of a young adult, including all the generalizations that arise from a fully developed visual hierarchy that can reason about occlusions and transparency. Standard RPMs are also restricted to being evaluated as a multiple-choice question, in contrast to the enactive evaluation in TW. In contrast to datasets that measure pixel accurate image reconstructions in simplistic settings \cite{higgins2017scan}, use of the tabletop world recognizes the schematic nature of concepts \cite{MandlerOnSchemas}, and enables the evaluation of sensorimotor representations for their generalization to settings that are different from the training distribution including different real world settings involving robots.

One common approach in program synthesis is to combine a domain specific language with a context-free grammar, such that the whole space of valid programs is derivable \cite{MandlerOnSchemas,Muggleton2015meta} (all programs that are generated would compile, and all programs that solve a task would be generated with non-zero probability). In order to achieve this, a popular choice is to use functional programming with type checking, which guarantees successful execution of programs generated according to the grammar. In contrast, we found that an imperative programming language was more suited for our purpose and subjectively more adequate to describe the thought process associated to a concept. This does not guarantee error-free execution when sampling from a Markov model, and some programs are rejected: the machine itself becomes part of the model, conditioning on valid programs (as in rejection sampling), and effectively pruning the search space. We took inspiration from the recent work \cite{ChenTOWARDSEXAMPLES} in bringing the machine itself into program synthesis.

We are excited about the future research directions that our work opens up. A richer set of primitive instructions that support an interplay of bottom-up and top-down attention, and utilizes the part-based representation of the vision hierarchy could enable a wider variety of concepts. Perhaps some of the primitives could be learned by having an agent interact with the environment using mechanisms we elaborated earlier \cite{Hay2018BehaviorContingencies}. The use of joint attention by pointing can be expanded in scope to direct top-down attention, not just foveation. Attributes of an object -- width, height, number of corners etc -- could themselves be represented as sensorimotor programs that are learned with experience in the real world, but evaluated purely in imagination during execution. Expanding the dynamics model to include three dimensional objects, and combining it with occlusion reasoning abilities of the vision hierarchy could result in a large number of real-world concepts being learned as cognitive programs.

\section*{Methods}

\subsection*{Program induction}

We are provided with a set of collections of input-output pairs of images, with each collection corresponding to some unknown concept, and we wish to infer the programs that describe each of those concepts. We refer to inferring such a program as ``discovering'' the concept. Knowing the program associated to a concept is equivalent to understanding it, since for new data (and without any additional supervision) we can now obtain the outputs from the inputs, and therefore also check for the presence of a concept.

A na\"ive way to perform this inference is via brute force. For each collection we could simply try every possible program until one of them explains every input-output transformation in the collection. This approach becomes unfeasible very quickly, and it is only useful to discover concepts that can be expressed as very short programs in our language of choice. It can be nonetheless useful as an initial step, since a brute force exploration will already provide us with the simplest concepts, which we can use as bootstrapping set.

To discover more sophisticated programs, we can fit a probabilistic model to the programs in the bootstrapping set, and then use that model to guide the search: start from the most probable program and then search in order of decreasing probability. Once some effort threshold is hit (for instance, number of programs considered), collect the found programs, re-fit the probabilistic model, and repeat. This approach is referred to as the explore-compress (EC) framework in \cite{Dechter2013}, where compression stands for fitting a probabilistic model to data.

The key for this approach to work properly is the probabilistic model. It has to be able to fit our programs well enough to avoid underfitting (thus exploring mostly uninteresting portions of the program space), but at the same time avoid overfitting (thus never departing from the already known set of training programs). Additionally, it has to allow exploring its domain in order of decreasing probability.

In order to achieve these desiderata we model programs (both instructions and arguments) as an observed Markov chain. The model for the instructions is learned from the already discovered concepts, whereas the emission model is conditional on the input-output pairs of examples and is learned separately. As we will see next, the whole induction process depends only on two free parameters: A modeling parameter $\varepsilon$ (the pseudocount) and an exploration parameter $M$ (maximum number of explored nodes).

\subsubsection*{The probabilistic model}

We start by considering programs as a sequence of instructions, without arguments. We define a program $x$ as a sequence of atomic instructions $x = [x_1, x_2,\ldots,x_L]$, where the last instruction is always $x_L=e$, a special end-of-program marker. This marker allows to sample finite programs of variable length from the Markov model. The probability of a single program is
$$
\log p(x) = \log p(x_1) + \sum_{i=1}^{L-1} \log p(x_{i+1}|x_i)
$$
where $p(x_{i+1}|x_i)$ is the transition probability from instruction $x_i$ to instruction $x_{i+1}$, and $p(x_1)$ are the initial probabilities. To compute the probability of multiple programs $\{x^{(i)}\}$, since we consider them independent, we can just add the log-probabilities of each of them. We can express this compactly by defining $X \equiv e~x^{(1)}~x^{(2)}~\ldots$ as a sequence that simply concatenates the programs (in any order) and prepends an end-of-program marker. Then, the joint probability of multiple sequences is simply
\begin{equation}
\label{eq:logprobinstr}
\log p(X) = \sum_{i=1}^{N-1} \log p(X_{i+1}|X_i)
\end{equation}
where $N$ is the total length of all the programs combined
\footnote{The starting probabilities are now expressed as transitioning probabilities from $e$. Also, note that since we know $X_1=e$ deterministically, $\log p(X_1) = 0$ and therefore it no longer appears in the expression.}, including the initial marker. From a compression perspective, $-\log p(X)$ corresponds to the description length (in nats) of the collection of programs under this model.

Parameter fitting for this model amounts to determining the transition matrix $T_x$, where $[T_x]_{rs} = p(s|r)$. The maximum likelihood estimator corresponds to simply counting the number of times a given instruction follows another in the data and normalizing. Simply doing this would however result in overfitting problems: if we have never observed a particular instruction transition, its probability would become zero and rule out such programs, preventing their future discovery. This problem can be avoided by additive smoothing\footnote{This is equivalent to placing a Dirichlet prior with parameter $1 + \varepsilon$ over the transition probabilities and obtaining their maximum a posteriori estimation.} with a small \emph{pseudocount} $\varepsilon$, which amounts to maximum a posteriori (MAP) inference.

As defined so far, the model is lacking any context beyond the previous instruction. We can further enhance it by using subroutines. Subroutines are sequences of instructions. We can augment the previous model with subroutines simply by adding a dictionary $D$ with their definitions, and allowing instructions to index not only atomic instructions, but also subroutines. A program with \emph{atomic} instructions $x$ can now be expressed in \emph{compressed} form $c$ by identifying the subroutines it contains and replacing them with a single instruction containing the appropriate subroutine call. There are potentially several ways to compress a program for a given set of subroutines. The reverse, decompressing a program $c$ into its atomic form $x$ is a deterministic process that always results into the original program. The joint probability of all the programs $X$, its compressed representations $C$, and the subroutine dictionary $D$ is then
\begin{equation}
\label{eq:logprobdict}
\log p(X, C, D) = \log p(X|C, D) + \log p(C) + \log p(D) = \log p(C) + \log p(D)
\end{equation}
where the last equality follows from $X$ being deterministically obtained from $C$ and $D$, and therefore $\log p(X|C, D)=0$ for valid expansions $X$ of $C$. The two terms in the r.h.s.\ can each be encoded as concatenated sequences (as we did for $X$) and computed using \eqref{eq:logprobinstr}.

In order to fit this model, we want to maximize \eqref{eq:logprobdict} w.r.t.\ the transition matrix $T_c$ (which is shared for both programs and subroutines), the compressed representation $C$, and the dictionary $D$. This joint optimization is in general a hard problem. We opt for a greedy method: we consider the increase in $\log p(X, C, D)$ that inserting each new possible subroutine in the dictionary (and updating $C$ and $T_c$ accordingly) would produce, and insert the one that achieves the maximum gain. Then we repeat the process until no positive increase is achievable, adding one subroutine at a time. We only consider as potential subroutines sequences of instructions that appear multiple times in the programs.

Observe that the maximum a posteriori (MAP) estimation of $D$, unlike the maximum likelihood one, includes the prior. Ignoring the prior probability of the dictionary subroutines would lead to failure here, since the optimal solution would be to simply move all the programs to the dictionary and encode each program as a single instruction. In contrast, MAP provides a trade-off in which subroutines are only deemed useful if they appear often enough.

We now turn to modeling the arguments of the instructions of a program.
Each program $x$ has an accompanying set of parameters $y = [y_1, y_2,\ldots,y_L]$ of the same length as $x$. Thus each instruction has exactly one parameter, which can take one value among a discrete set of choices. Those choices are different for each instruction and are given by the syntax of the language. For a given dictionary, the probability of a full program including now arguments is
\begin{align}
\label{eq:logprobarguments}
\nonumber \log p(y, x, c | D) & = \log p(x|c, D) + \sum_{j=1}^{L_c-1} \log p(c_{j+1}|c_j)  + \sum_{i=1}^{L-1} \log p(y_{i+1}|x_{i+1}) \\
\nonumber &=\log p(x|c, D) + \sum_{j=1}^{L_c-1} \left[\sum_{h=1}^{n_{j+1}} \log p(y_{j+1h}|x_{j+1h}) + \log p(c_{j+1}|c_j)\right]\\
&=\sum_{j=1}^{L_c-1} \log p(y_{j+1}|c_{j+1}) + \log p(c_{j+1}|c_j)
\end{align}
where we have collected all the arguments that are used in a subroutine $c_j$ into a single variable $y_j=[y_{j1},\ldots,y_{jn_j}]$ and removed $\log p(x|c, D) = 0$ (due to determinism) from the equation.
We have already estimated all the quantities in the above expression except for the conditional probability of the arguments $\log p(y_i|x_i)$ of a given atomic instruction\footnote{As described in the equation, the conditional probability of the compounded arguments of a subroutine can be obtained simply by multiplying the conditional probabilities of the involved atoms.}. Those can be estimated in multiple ways: one can consider a uniform probability over arguments, or a more informed one based on data (see section on argument prediction).

\subsubsection*{The exploration}

We are now interested in discovering the concept that explains a particular input-output collection.
It is useful to first express our model over programs with arguments \eqref{eq:logprobarguments} as a simple Markov chain with no emissions. To that end, we simply combine the compressed states $c_j$ and their compressed arguments $y_j$ into a single joint variable $z_j \equiv (c_j, y_j)$. Now the probability of a compressed program $z$ can be expressed as a simple Markov chain
$$
\log p(z) = \sum_{j=1}^{L_c-1} \log p(z_{j+1}|z_j)
$$
whose transition probability $T_z$ can be easily derived by combining $T_c$ and $p(y_j|c_j)$.

We thus have a Markov model over ``program portions'' $z_j$. Those program portions can be either individual instructions or subroutines (both fully parameterized). We want to sample full programs from this model in order of decreasing probability (i.e., increasing description lengths). To do this, we notice that the Markov model induces an exploration tree: each node has as children all the possible program portions, and is connected to them through arcs with cost $-\log p(z_\text{child}|z_\text{parent})$. The root is an end-of-program marker, and each node corresponds to a program that can be read off by traversing the path from the root to the node. The description length of such program can be obtained by adding the weights of the arcs on the path.

A best-first search traversal of the tree (always expanding the nodes that have less accumulated cost from the root) results in visiting the programs in order of decreasing probability. Then we can expand each visited compressed program $z$ into its atomic version $x$ and run it in our emulator, checking whether the produced outputs match the provided ones. We stop the process after we find a valid program (the concept is discovered), or the number of visited nodes exceeds some effort parameter $M$.

In practice, this process can be very memory demanding, since in the worst case it requires to store all the nodes of the tree. To avoid this problem, iterative deepening can be used. In iterative deepening, depth-first search (DFS) is run with a limit on the \emph{description length}. Then this limit is increased and the process repeated. The nodes (programs) visited on each iteration that were not visited in the previous iteration are then run on the emulator. This visits the nodes in approximately best first order: within each search bracket the node ordering is arbitrary, but the brackets are ordered. The smaller that the brackets are, the tighter the approximation to best-first search is.

Two additional optimizations are feasible: Each node that we visit contains a program that will correspond to a previously run program plus some additional instructions. Since within each bracket we are following a DFS traversal, it is feasible to keep a copy of the emulators that were run to obtain each of the prefixes of the current program and use them to execute only the additional instructions. Furthermore, if the emulator deems a program to be ``invalid'', then we can prune the search from that point down (since all programs with that prefix will also be invalid), which is trivial in a DFS traversal.

In our case, since the number of explored programs is relatively low, we first run a best-first search that identifies the maximum description length of the shortest $M$ programs (without running them), and then we run DFS using the identified description length as the cut-off. During the DFS we do run the programs, taking advantage of the optimizations described in the previous paragraph. The DFS is run until completion even if the sought-for concept is found earlier, since it does not guarantee that the shortest description of the concept will be found first.

\subsection*{Argument prediction}
\label{sec:argprediction}
The differences and similarities between the input and output images that are used as examples for each concept provide information about what to pay attention to. We use this information to predict the argument for the following three functions: {\tt set\_shape\_attn}, {\tt set\_color\_attn}, {\tt fill\_color}. Given that the argument of each function can take several potential values and each function can appear multiple times in a program with its argument taking a different value, this becomes a multi-label problem. We build a logistic regression model for each function and argument pair.

In order to get training data for argument prediction, we enumerate all valid programs up to length 6 (including {\tt scene\_parse}), where valid programs are those that can be executed on input images without any failure. From the set of valid programs, we filter out uninformative programs whose output images are the same as the corresponding input. For example, a program [{\tt scene\_parse()}, {\tt set\_color\_attn(`r')}, {\tt top\_down\_attn()}] has the same output images as its input. We also remove programs where an instruction with its assigned argument is not indispensable. For example, {\tt set\_shape\_attn(`square\_shape')} is redundant for the program [{\tt scene\_parse()}, {\tt set\_shape\_attn(`square\_shape')}, {\tt set\_shape\_attn(`circle\_shape'}), {\tt top\_down\_attn()}, {\tt fill\_color(`red')}]. The criteria for a function with its assigned argument to be indispensable: if the function and argument pair is removed from a program, the program would produce a different output.

For each program, we generate 10 examples. Each example is converted to a 3D binary array $A$ with shape (21, 15, 15), where 21 is the number of input channels and 15 corresponds to the height and width of the discretized images. The first ten channels are based on the input images, as shown in the first column of Supplementary Fig S2. Each element is set to 1 or 0, depending on whether the feature associated to that channel is present at that location or not, respectively. The next ten channels are based on the difference between input and output images, both using the same binary encoding, which results in the elements of these channels having as possible values -1, 0, and 1. This is shown in the third column of Supplementary Fig S2. The difference between output and input highlights what has changed between both images. For example, when the green color disappears, a -1 value is registered in the third column at the corresponding location and for the green color channel. In contrast, the blue square remains in the same position, thus it becomes zero when we subtract the input from the output.
The last channel summarizes the differences across the ten previous features (channels). Thus element corresponding to row $r$ and column $c$ of channel 21 is computed as $A_{21rc} = \sum_{f=11}^{20} |A_{frc}| \geq 1$. I.e., a value of 1 at a given position means that the object at that position has changed regardless of whether it was added or removed. This feature combines color change and movement change into one single indicator.

%\begin{figure}
%    \centering
%    \includegraphics[width=4in]{arg_pred_features.pdf}
%    \caption{Features for argument prediction}
%    \label{arg_pred_features}
%\end{figure}

We use a Convolutional Neural Network (CNN) to capture spatial invariance, since the object experiencing a change can be anywhere in an image. Supplementary Fig S3 shows the architecture of the model. In order to capture the similarities among examples, we sum the max-pooling results from all examples and feed it into a sigmoid function. Given that the last layer is simply performing a summation, we only need to train the convolutional weights of the CNN. The model is trained using the ADAM optimizer and L1 regularization with 0.01 weight.

%\begin{figure}
%    \centering
%    \includegraphics[width=6in]{fig7.pdf}
%    \caption{CNN architecture for argument prediction}
%    \label{arg_pred_arch}
%\end{figure}

There are other two functions with arguments that are not supported by this model and for which we do not perform predictions, one of them is {\tt foveate} and the other one is {\tt imagine\_shape}.

We also use the result of argument prediction to predict the existence of an instruction in a concept. Specifically, if for a given instruction the sum of probabilities of all the values that its argument can take is below 0.55, it is assumed that the instruction is not present in the concept and excluded from the search process.

\subsection*{Execution on Robots}
\label{sec:robotexec}
We tested the transfer of induced concepts to real world by executing programs on two different robots in different settings: a Baxter robot from Rethink Robotics, and a UR5 from Universal Robots. To execute the programs on these robots, we extended our emulator with an additional robot interface that implemented input scene capture and hand actions. Scene capture is achieved through a camera attached to the respective robot's end effector, a gripper. A rgb image of the scene is captured by this camera and passed on to the scene parser of the emulator, which creates the emulator's initial state. Moving hand to any location within the workspace, and grasping and releasing objects are implemented on robot using a simple Cartesian controller that moved the hand to a given x, y position on a table in a plane at a specified height above the table top. Grasping and releasing an object involved moving the gripper down and closing and opening the gripper. Instead of dragging the object along the table, we moved the object slightly above the table but otherwise respected the same collision constraints as our table top world including the boundary. Interface also maps any position in emulator's workspace onto x, y coordinates in the robot frame of reference. Program execution took place in the emulator and called scene capture and hand action functions implemented in the robot interface when available. The rest of the VCC instructions remained the same as our emulator. Executing any specific program involved giving the induced concept as a list of primitives with arguments when available and running the emulator with robot interface. We used the same visual scene parser for execution on robots as we did for the emulator. We tested 6 different concepts, including a complex concept that involved executing two concepts in sequence. We use colored foam blocks of different shapes, fruits, and household items as objects and executed programs under different variations of background, number, shapes and types of objects, and with or without distractor objects.

Mapping locations from emulator workspace to robot reference frame requires accurate calibration of camera pose with respect to the robot, and moving robot arm to a specific location requires accurate execution of arm movement. Our UR5 robot had an external realsense camera attached to the gripper with an accurate calibration. For Baxter, an inbuilt camera inside the gripper is used instead for scene capture. This camera is of low-resolution and has some burnout pixels with an approximate calibration. Due to these limitations, executions are typically accurate and more successful on UR5 compared to Baxter. Movement execution is also faster on UR5 compared to Baxter.

In our tests, failed runs of program execution are primarily due to grasp failures. These happened more frequently on Baxter, so we tested most of the variations on UR5 which has better movement accuracy and better calibration of camera with respect to the robot.

\clearpage

% Your references go at the end of the main text, and before the
% figures.  For this document we've used BibTeX, the .bib file
% scibib.bib, and the .bst file Science.bst.  The package scicite.sty
% was included to format the reference numbers according to *Science*
% style.

% \bibliography{mendeley_v2}
\bibliography{mendeley_fixed_2}

\bibliographystyle{science}

% Following is a new environment, {scilastnote}, that's defined in the
% preamble and that allows authors to add a reference at the end of the
% list that's not signaled in the text; such references are used in
% *Science* for acknowledgments of funding, help, etc.

%\begin{scilastnote}
%\item We've included in the template file \texttt{scifile.tex} a new
%environment, \texttt{\{scilastnote\}}, that generates a numbered final
%citation without a corresponding signal in the text.  This environment
%can be used to generate a final numbered reference containing
%acknowledgments, sources of funding, and the like, per {\it Science\/}
%style.
%\end{scilastnote}

% For your review copy (i.e., the file you initially send in for
% evaluation), you can use the {figure} environment and the
% \includegraphics command to stream your figures into the text, placing
% all figures at the end.  For the final, revised manuscript for
% acceptance and production, however, PostScript or other graphics
% should not be streamed into your compliled file.  Instead, set
% captions as simple paragraphs (with a \noindent tag), setting them
% off from the rest of the text with a \clearpage as shown  below, and
% submit figures as separate files according to the Art Department's
% instructions.

\clearpage

%\noindent {\bf Fig. 1.} Please do not use figure environments to set
%up your figures in the final (post-peer-review) draft, do not include graphics in your
%source code, and do not cite figures in the text using \LaTeX\
%\verb+\ref+ commands.  Instead, simply refer to the figure numbers in
%the text per {\it Science\/} style, and include the list of captions at
%the end of the document, coded as ordinary paragraphs as shown in the
%\texttt{scifile.tex} template file.  Your actual figure files should
%be submitted separately.

\end{document}